\documentclass{article}
\usepackage[utf8]{inputenc}
\usepackage{graphicx}
\usepackage{url}
\usepackage{float}
\usepackage{caption}
\usepackage{hyperref}
\usepackage{multirow}
\usepackage{placeins}
\usepackage{natbib}
\setcitestyle{authoryear,open={(},close={)}} 
\usepackage[margin=1.2in]{geometry}
\title{ARAGOG: Advanced RAG Output Grading}
\author{Matouš Eibich \\ Predli \\ \href{matous@predli.com}{matous@predli.com} 
   \and Shivay Nagpal \\ Predli \\ \href{shivay@predli.com}{shivay@predli.com} 
   \and Alexander Fred-Ojala \\ Predli \& UC Berkeley \\ \href{afo@berkeley.edu}{afo@berkeley.edu} }

\begin{document}
\maketitle

\begin{abstract}
Retrieval-Augmented Generation (RAG) is essential for integrating external knowledge into Large Language Model (LLM) outputs. While the literature on RAG is growing, it primarily focuses on systematic reviews and comparisons of new state-of-the-art (SoTA) techniques against their predecessors, with a gap in extensive experimental comparisons. This study begins to address this gap by assessing various RAG methods' impacts on retrieval precision and answer similarity. We found that Hypothetical Document Embedding (HyDE) and LLM reranking significantly enhance retrieval precision. However, Maximal Marginal Relevance (MMR) and Cohere rerank did not exhibit notable advantages over a baseline Naive RAG system, and Multi-query approaches underperformed. Sentence Window Retrieval emerged as the most effective for retrieval precision, despite its variable performance on answer similarity. The study confirms the potential of the Document Summary Index as a competent retrieval approach. All resources related to this research are publicly accessible for further investigation through our GitHub repository \href{https://github.com/predlico/ARAGOG}{ARAGOG}. We welcome the community to further this exploratory study in RAG systems.
\end{abstract}

\section{Introduction}
\begin{figure}[htbp]
    \centering
    \captionsetup{width=0.95\linewidth, font=footnotesize} 
    \includegraphics[width=0.95\linewidth]{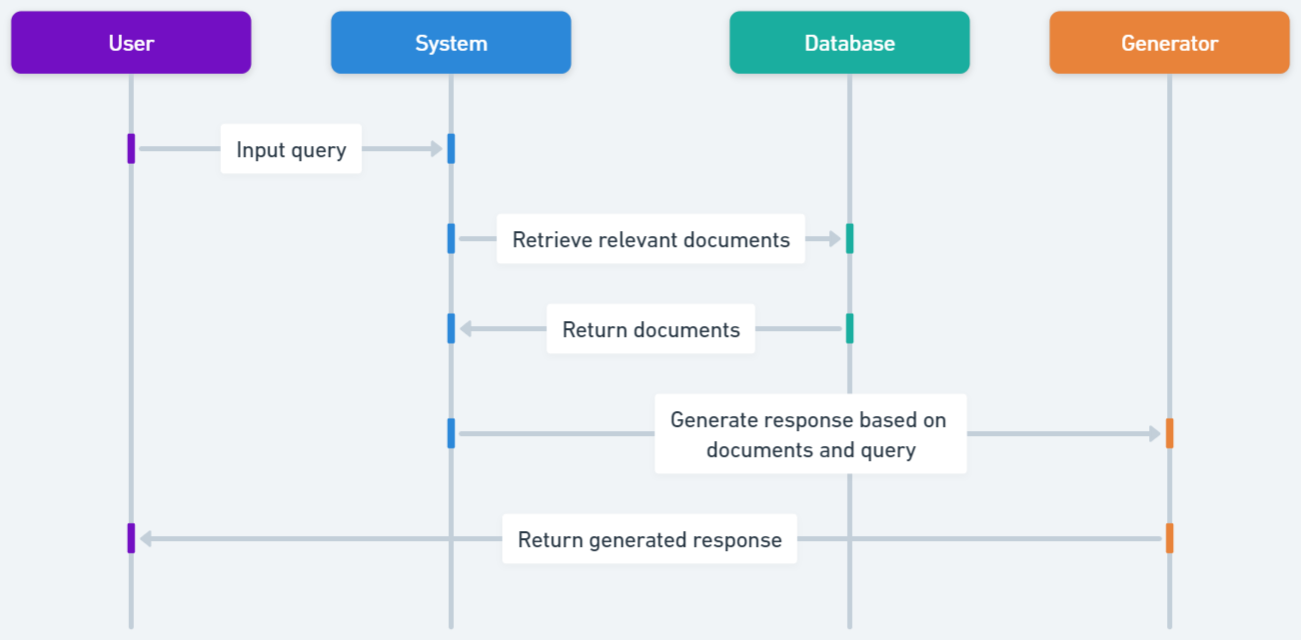}
    \caption[Workflow of a RAG System]{A high-level overview of the workflow within a Retrieval-Augmented Generation (RAG) system. This process diagram shows how a user query is processed by the system to retrieve relevant documents from a database and how these documents inform the generation of a response.}
    \label{fig:rag-system-workflow}
\end{figure}

Large Language Models (LLMs) have significantly advanced the field of natural language processing, enabling a wide range of applications from text generation to question answering. However, integrating dynamic, external information remains a challenge for these models. Retrieval Augmented Generation (RAG) techniques address this limitation by incorporating external knowledge sources into the generation process, thus enhancing the models' ability to produce contextually relevant and informed outputs. This integration of retrieval mechanisms with generative models is a key development in improving the performance and versatility of LLMs, facilitating more accurate and context-aware responses. See Figure~\ref{fig:rag-system-workflow} for an overview of the standard RAG workflow.

Despite the growing interest in RAG techniques within the domain of LLMs, the existing body of literature primarily consists of systematic reviews \citep{gao2024retrievalaugmented} and direct comparisons between successive state-of-the-art (SoTA) models \citep{gao2022precise, jiang2023active}. This pattern reveals a notable gap: a comprehensive experimental comparison across a broad spectrum of advanced RAG techniques is missing. Such a comparison is crucial for understanding the relative strengths and weaknesses of these techniques in enhancing LLMs' performance across various tasks. This study seeks to contribute to bridging this gap by providing an extensive evaluation of multiple RAG techniques and their combinations, thereby offering insights into their efficacy and applicability in real-world scenarios.

The focus of this investigation is a spectrum of advanced RAG techniques aimed at optimizing the retrieval process. These techniques can be categorized into several areas:

\begin{table}[ht]
\centering
\begin{tabular}{|c|c|}
\hline
\textbf{RAG Technique} & \textbf{Type}\\ 
\hline
Sentence-window retrieval & \multirow{2}{*}{Decoupling of Retrieval and Generation} \\ \cline{1-1}
Document summary index & \\ 
\hline
HyDE & \multirow{2}{*}{Query Expansion}                        \\ \cline{1-1}
Multi-query                      &                                                         \\ \hline
Maximal Marginal Relevance (MMR) & Enhancement Mechanism                                   \\ \hline
Cohere Re-ranker                & \multirow{2}{*}{Re-rankers}                             \\ \cline{1-1}
LLM-based Re-ranker              &                                                         \\ \hline
\end{tabular}
\end{table}

\FloatBarrier

To evaluate the RAG techniques, this study leverages two metrics: Retrieval Precision and Answer Similarity \citep{tonic2023ragmetrics}. Retrieval Precision measures the relevance of the retrieved context to the question asked, while Answer Similarity assesses how closely the system's answers align with reference responses, on a scale from 0 to 5. 

\section{RAG Techniques}

\subsection{Sentence-window retrieval}

The Sentence-window Retrieval technique is grounded in the principle of optimizing both retrieval and generation processes by tailoring the text chunk size to the specific needs of each stage \citep{towardsdatascience2023advancedrag}. For retrieval, this technique emphasizes single sentences to take advantage of small data chunks for potentially better retrieving capabilities. On the generation side, it adds more sentences around the initial one to offer the LLM extended context, aiming for richer, more detailed outputs. This decoupling is supposed to increase the performance of both retrieval and generation, ultimately leading to better performance of the whole RAG system.

\subsection{Document summary index}

The Document Summary Index method enhances RAG systems by indexing document summaries for efficient retrieval, while providing LLMs with full text documents for response generation \citep{llamaindex2023docsummary}. This decoupling strategy optimizes retrieval speed and accuracy through summary-based indexing and supports comprehensive response synthesis by utilizing the original text.

\subsection{HyDE}

\begin{figure}[htbp]
    \centering
    \includegraphics[width=0.95\linewidth]{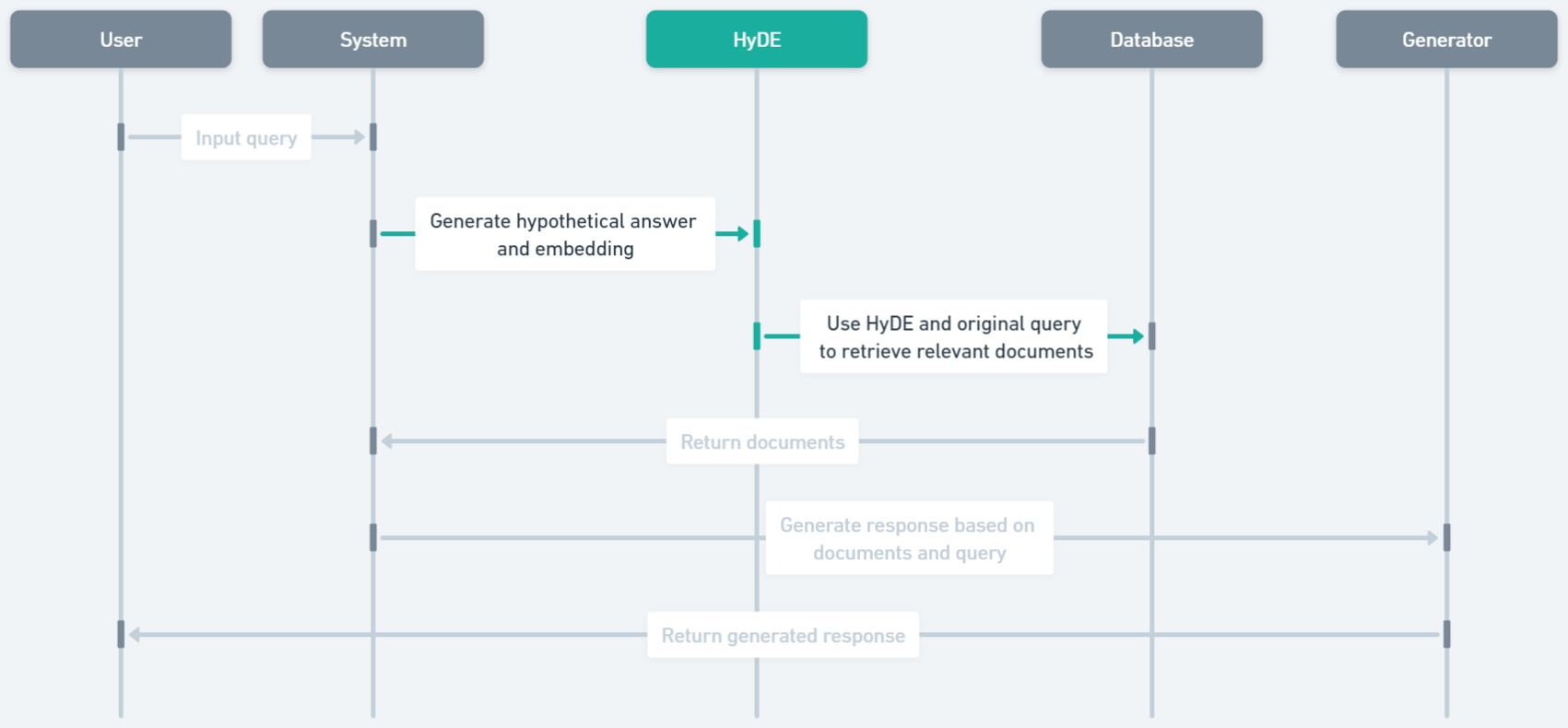}
    \captionsetup{width=0.95\linewidth, font=footnotesize} 
    \caption[HyDE Process Flow]{The process flow of Hypothetical Document Embedding (HyDE) technique within a Retrieval-Augmented Generation system. The diagram illustrates the steps from the initial query input to the generation of a hypothetical answer and its use in retrieving relevant documents to inform the final generated response.}
    \label{fig:hyde-process-flow}
\end{figure}

The Hypothetical Document Embedding \citep{gao2022precise} technique enhances the document retrieval by leveraging LLMs to generate a hypothetical answer to a query. HyDE capitalizes on the ability of LLMs to produce context-rich answers, which, once embedded, serve as a powerful tool to refine and focus document retrieval efforts. See Figure~\ref{fig:hyde-process-flow} for overview of HyDE RAG system workflow. 

\subsection{Multi-query}

\begin{figure}[htbp]
    \centering
    \includegraphics[width=0.95\linewidth]{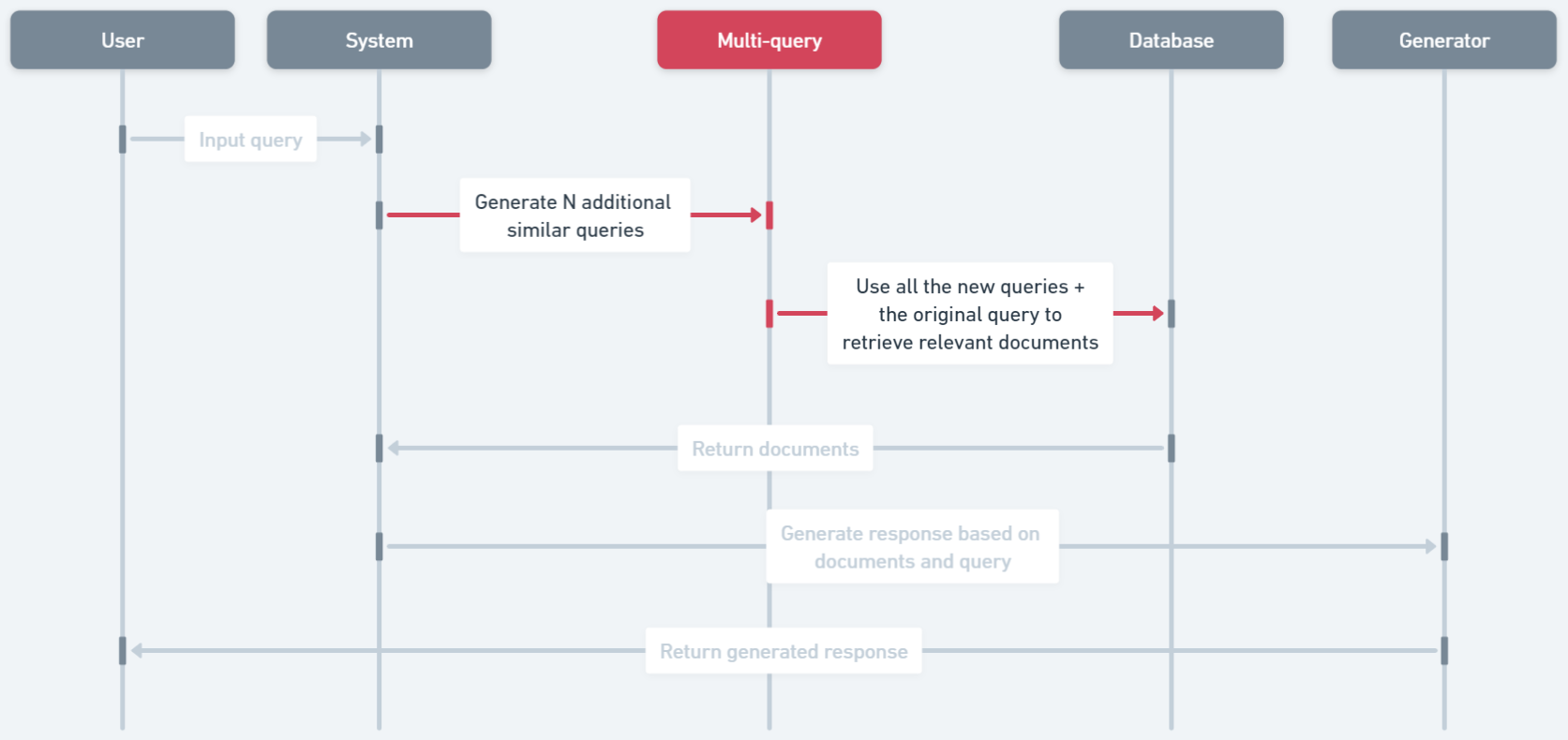}
    \captionsetup{width=0.95\linewidth, font=footnotesize} 
    \caption[Multi-query Enhancement Flow]{This diagram showcases how multiple similar queries are generated from an initial user query, and how they contribute to retrieving a wider range of relevant documents.}
    \label{fig:multi-query-enhancement-flow}
\end{figure}

The Multi-query technique \citep{langchain_blog2023query} enhances document retrieval by expanding a single user query into multiple similar queries with the assistance of an LLM. This process involves generating \textit{N} alternative questions that echo the intent of the original query but from different angles, thereby capturing a broader spectrum of potential answers. Each query, including the original, is then vectorized and subjected to its own retrieval process, which increases the chances of fetching a higher volume of relevant information from the document repository. To manage the resultant expanded dataset, a re-ranker is often employed, utilizing machine learning models to sift through the retrieved chunks and prioritize those most relevant in regards to the initial query. See Figure~\ref{fig:multi-query-enhancement-flow} for an overview of how Multi-query RAG system workflow. 

\subsection{Maximum Marginal Relevance}

The Maximal Marginal Relevance (MMR) technique \citep{carbonell1998usemmr} aims to refine the retrieval process by striking a balance between relevance and diversity in the documents retrieved. By employing MMR, the retrieval system evaluates potential documents not only for their closeness to the query's intent but also for their uniqueness compared to documents already selected. This approach mitigates the issue of redundancy, ensuring that the set of retrieved documents covers a broader range of information. 

\subsection{Cohere Rerank}

\begin{figure}[htbp]
    \centering
    \includegraphics[width=0.95\linewidth]{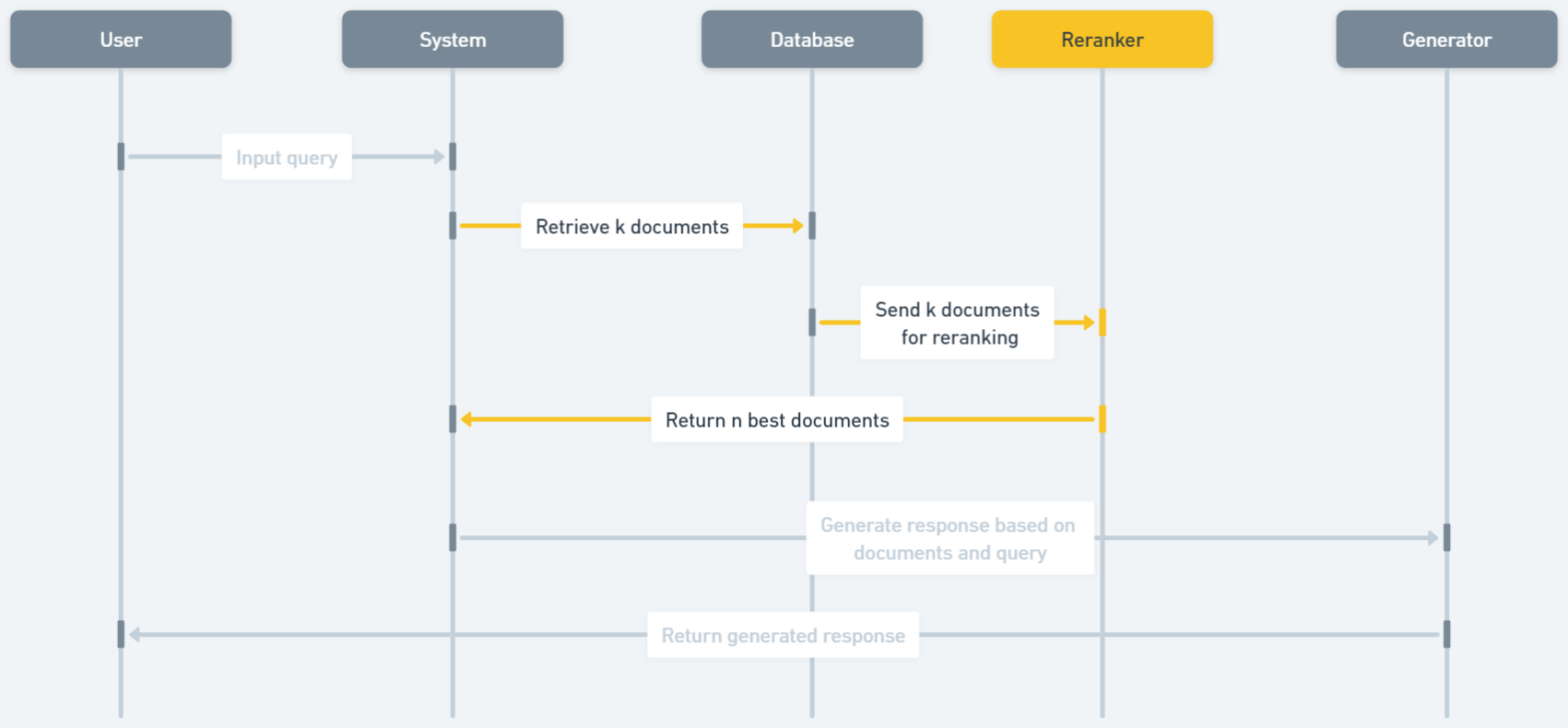}
    \captionsetup{width=0.95\linewidth, font=footnotesize} 
    \caption[Reranking Process Flow]{This flowchart outlines the reranking process in a RAG system. It illustrates how retrieved documents are further assessed for relevance using a reranking step, which refines the set of documents that will inform the generated response.}
    \label{fig:reranking-process-flow}
\end{figure}

Rerankers aim to enhance the RAG process by refining the selection of documents retrieved in response to a query, with the goal of prioritizing the most relevant and contextually appropriate information for generating responses \citep{pinecone2023rerankers}. This step employs ML algorithms (such as cross-encoder) to reassess the initially retrieved set, using criteria that extend beyond cosine similarity. Through this evaluation, rerankers are expected to improve the input for generative models, potentially leading to more accurate and contextually rich outputs. See Figure~\ref{fig:reranking-process-flow} for an overview of the Reranker RAG system workflow.

One tool in this domain is Cohere rerank, which uses a cross-encoder architecture to assess the relevance of documents to the query. This approach differs from methods that process queries and documents separately, as cross-encoders analyze them jointly, which could allow for a more comprehensive understanding of their mutual relevance.

\subsection{LLM rerank}

Following the introduction of cross-encoder based rerankers such as Cohere rerank, the LLM reranker offers an alternative strategy by directly applying LLMs to the task of reranking retrieved documents \citep{llamaindex2023llmrerank}. This method prioritizes the comprehensive analytical abilities of LLMs over the joint query-document analysis typical of cross-encoders. Although less efficient in terms of processing speed and cost compared to cross-encoder models, LLM rerankers can achieve higher accuracy by leveraging the advanced understanding of language and context inherent in LLMs. This makes the LLM reranker suitable for applications where the quality of the reranked results is more critical than computational efficiency. Figure~\ref{fig:reranking-process-flow} workflow applies to LLM reranker as well.

\section{Methods}
\subsection{Data}
This study utilizes a tailored dataset derived from the \textit{AI ArXiv} collection, accessible via Hugging Face \citep{calam2023aiarxiv}. The dataset consists of 423 selected research papers centered around the themes of AI and LLMs, sourced from arXiv. This selection offers a comprehensive foundation for constructing a database to test the RAG techniques and creating a set of evaluation data to assess their effectiveness.

\subsubsection{RAG Database Construction}

For the study, a subset of 13 key research papers was selected for their potential to generate specific, technical questions suitable for evaluating Retrieval-Augmented Generation (RAG) systems. Among the selected papers were significant contributions such as \textit{RoBERTa: A Robustly Optimized BERT Pretraining Approach} \citep{liu2019roberta} and \textit{BERT: Pre-training of Deep Bidirectional Transformers for Language Understanding} \citep{devlin2019bert}. To better simulate a real-world vector database environment, where noise and irrelevant documents are present, the database was expanded to include the full dataset of 423 papers available. The additional 410 papers act as noise, enhancing the complexity and diversity of the retrieval challenges faced by the RAG system. 

\subsubsection{Chunking Approach}
Multiple chunking strategies were utilized to create vector databases for different retrieval methods. For the classic vector database, a \href{https://docs.llamaindex.ai/en/stable/api_reference/node_parsers/token_text_splitter/}{\textit{TokenTextSplitter}} was employed with a chunk size of 512 tokens and an overlap of 50 tokens. This approach split the documents into smaller chunks while maintaining context by allowing for overlapping text between chunks. For the sentence window method, a \href{https://docs.llamaindex.ai/en/stable/examples/node_postprocessor/MetadataReplacementDemo/}{\textit{SentenceWindowNodeParser}} was used with a window size of 3 sentences, effectively creating overlapping chunks consisting of three consecutive sentences. Lastly, for the document summary index, a \href{https://docs.llamaindex.ai/en/stable/api_reference/node_parsers/token_text_splitter/}{\textit{TokenTextSplitter}} was employed with a larger chunk size of 3072 tokens and an overlap of 100 tokens, generating larger chunks to be summarized by the language model.

\subsubsection{Evaluation Data Preparation}
\begin{figure}[htbp]
    \centering
    \includegraphics[width=0.95\linewidth]{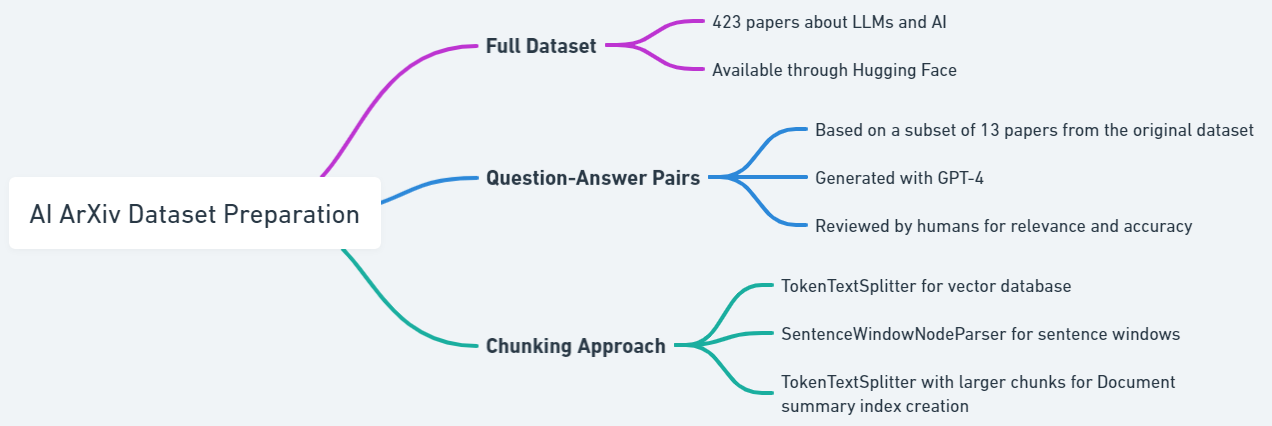}
    \captionsetup{width=0.95\linewidth, font=footnotesize} 
    \caption[AI ArXiv Dataset Preparation Process]{The visualization of the \textit{AI ArXiv} dataset preparation process. This diagram shows the selection of papers for question-answer generation, the employment of the full dataset to provide ample noise for the RAG system, and the chunking approaches used to process the documents for the vector database.}
    \label{fig:ai-arxiv-dataset-preparation}
\end{figure}

The evaluation dataset comprises 107 question-answer (QA) pairs generated with the assistance of GPT-4. The generation process was guided by specific criteria to ensure that the questions were challenging, technically precise, and reflective of potential user inquiries sent to a RAG system. Each QA pair was then reviewed by humans to validate its relevance and accuracy, ensuring that the evaluation data accurately measures the RAG techniques' performance in real-world applications. The QA dataset is available in this paper's associated Github repository \href{https://github.com/predlico/ARAGOG}{ARAGOG}. See Figure~\ref{fig:ai-arxiv-dataset-preparation} for an overview of the data preparation process.

\subsection{Mitigating LLM Output Variability} \label{sec-mitigation}
To address the inherent variability of LLM outputs, the methodology included conducting 10 runs for each RAG technique. This strategy was chosen to balance the need for statistical reliability against the limitations of computational resources and time. While more runs could increase statistical reliability, they would also complicate the need to distinguish between statistical and practical significance. 

\subsection{Metrics}
To evaluate the performance of various RAG techniques within this study, two primary metrics were employed from the Tonic Validate package/platform: Retrieval Precision and Answer Similarity \citep{tonic2023ragmetrics}. These metrics were selected to evaluate both the retrieval process and the generative capabilities of the LLMs used, with a primary focus on the precision of information retrieval.

\subsubsection{Retrieval Precision}
\begin{figure}[htbp]
    \centering
    \includegraphics[width=0.95\linewidth]{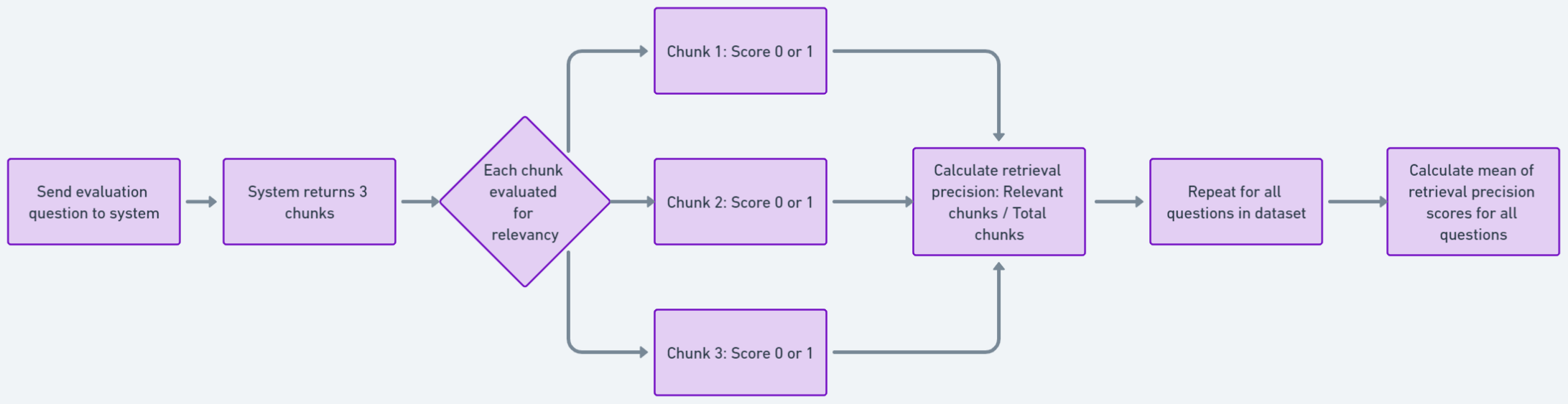}
    \captionsetup{width=0.95\linewidth, font=footnotesize} 
    \caption[Evaluation Process in RAG System]{This diagram illustrates the evaluation process in a RAG system, where each chunk's relevance is scored, contributing to the overall retrieval precision metric. The process highlights how individual chunks are evaluated for their utility in responding to user queries.}
    \label{fig:evaluation-process-rag}
\end{figure}

This metric serves as the cornerstone of the evaluation, directly measuring the efficacy of the retrieval techniques implemented in the RAG system. Retrieval Precision quantifies the percentage of context retrieved by the system that is relevant to answering a given question, with scores ranging from 0 to 1. A higher score indicates a greater proportion of the retrieved content is pertinent to the query. The evaluation is conducted by asking an LLM evaluator to determine the relevance of each piece of retrieved context, ensuring that the assessment focuses on the accuracy of information retrieval, rather than the subsequent generation quality. See Figure~\ref{fig:evaluation-process-rag} for visual explanation of retrieval precision grading workflow.

\subsubsection{Answer Similarity}
As a complementary metric, Answer Similarity assesses how well the answers generated by the RAG system align with reference answers, scored on a scale from 0 to 5. While this end-to-end test provides valuable insights into the system's overall performance, it was considered secondary to the primary objective of evaluating retrieval techniques. This is because Answer Similarity could be influenced by the generative capabilities of the LLM, potentially confounding the assessment of retrieval effectiveness alone.

\subsubsection{Rationale for Metric Selection}
To select appropriate metrics for the study the goal was to move beyond simplistic measures of similarity, such as cosine similarity between embeddings (which do not fully capture the complexity of effective retrieval and generation). The landscape of available metrics and evaluation platforms revealed a lack of consensus on optimal evaluation strategies for RAG systems, particularly with a focus on retrieval. While some methods, such as those proposed by RAGAS \citep{ragas2023metrics}, involve detailed calculations with LLMs and F1 scores, these were deemed to be unsuitable for the objectives in this paper. Their complexity often led to results that were unreliable. Conversely, simple embedding comparisons were deemed insufficient for capturing the nuanced effectiveness of retrieval techniques. Ultimately, the selection of Retrieval Precision as the primary metric, complemented by Answer Similarity, was driven by the focus on evaluating the retrieval component of RAG systems. Though confident in the appropriateness of these metrics, especially Retrieval Precision, one has to acknowledge the ongoing development in this area and remain open to future advancements and consensus in evaluation methodologies.

\subsection{LLM}
For the experimental setup GPT-3.5-turbo was selected because of  its cost-effectiveness and ease of implementation. Tonic Validate's requirement for OpenAI models led us to choose GPT-3.5-turbo due to its cost-effectiveness, even though GPT-4 might provide more precise grading at a higher expense. It is important to note that the choice of the LLM used for generation itself was less critical for the main objective—evaluating retrieval precision, since answer similarity can be regarded as a supplementary metric. 

\section{Results}
The study systematically evaluates a variety of advanced RAG techniques using metrics of Retrieval Precision and Answer Similarity. A comparative analysis is presented through boxplots to visualize the distribution of these metrics, followed by ANOVA and Tukey's HSD tests to determine the statistical significance of the differences observed. 
\subsection{Comparative Performance Analysis: Boxplots}
\begin{figure}[H]
    \centering
    \includegraphics[width=0.955\linewidth]{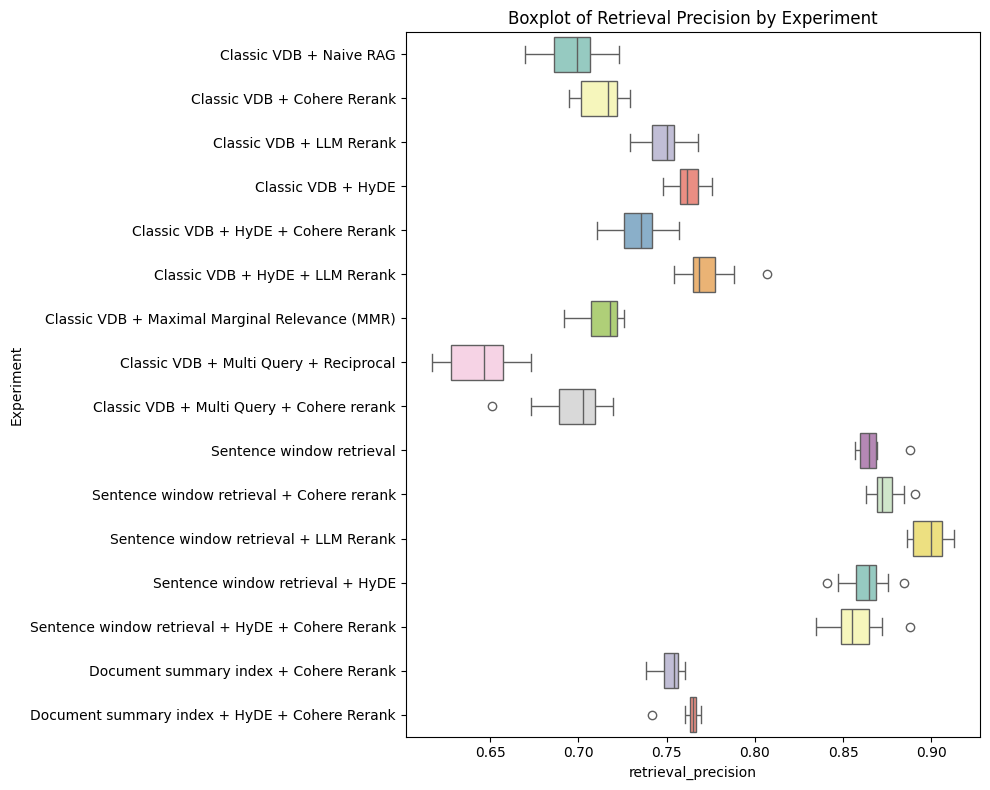}
    \captionsetup{width=.9\linewidth, font=footnotesize} 
    \caption[Boxplot of Retrieval Precision]{Boxplot of Retrieval Precision by Experiment. Each boxplot demonstrates the range and distribution of retrieval precision scores across different RAG techniques. Higher median values and tighter interquartile ranges suggest better performance and consistency.}
    \label{fig:retrieval-precision-boxplot}
\end{figure}

\begin{figure}[htbp]
    \centering
    \includegraphics[width=.9\linewidth]{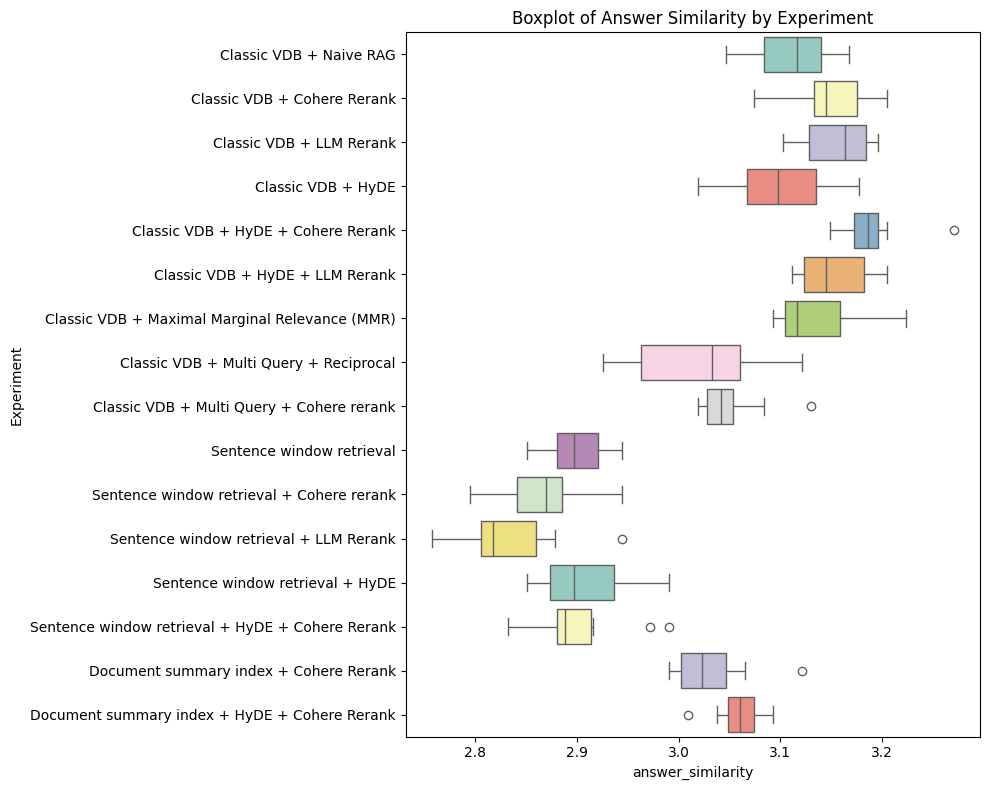}
    \captionsetup{width=.9\linewidth, font=footnotesize} 
    \caption[Boxplot of Answer Similarity]{Boxplot of Answer Similarity by Experiment. Each boxplot demonstrates the range and distribution of answer similarity scores across different RAG techniques. Higher median values and tighter interquartile ranges suggest better performance and consistency.}
    \label{fig:answer-similarity-boxplot}
\end{figure}

The boxplots for Retrieval Precision (Figure~\ref{fig:retrieval-precision-boxplot}) indicate varied performance across RAG techniques. The Sentence Window Retrieval approach is notably effective, with a high median precision, though this does not directly correlate with Answer Similarity performance (see Figure~\ref{fig:answer-similarity-boxplot}). Techniques utilizing LLM Rerank and Hypothetical Document Embedding (HyDE) show enhanced precision, markedly outperforming the Naive RAG baseline. Conversely, Maximal Marginal Relevance (MMR) and Cohere Rerank demonstrate limited benefits, with median precision scores comparable to or below the baseline. Multi-query, interestingly, presents a reduction in retrieval precision compared to Naive RAG, warranting further investigation into its application. Document summary index performance is similar to the best setting of Classic VDB, indicating that with further enhancements, Document summary technique would surpass Classic VDB. 

The analysis of answer similarity (Figure~\ref{fig:answer-similarity-boxplot}) presents intriguing patterns that both align with and diverge from those observed in retrieval precision. For Classic Vector Database (VDB) techniques and Document Summary Index, there is a notable positive correlation between retrieval precision and answer similarity, suggesting that when relevant information is accurately retrieved, it can lead to answers that more closely mirror reference responses.

In contrast, Sentence Window Retrieval displays a disparity between high retrieval precision and lower answer similarity scores. This could indicate that while the technique is adept at identifying relevant passages, it may not be translating this information into answers that are semantically parallel to the reference, possibly due to the generation phase not fully leveraging the retrieved context.

\subsection{Statistical Validation of Differences}
As described in section \ref{sec-mitigation}, 10 iterations of each experiment were conducted to mitigate the impact of inherent LLM variability on the results. An ANOVA test applied to these results confirmed significant differences in Retrieval Precision across the various techniques, validating that observed variations were not due to random chance but reflect true performance disparities. Following this, Tukey's Honestly Significant Difference (HSD) test provided a more granular understanding of these performance differentials. The statistical tests focus only on the primary metric of this study, retrieval precision. In light of the extensive range of possible pairwise comparisons, the analysis concentrated on the comparisons deemed to be most relevant.

\subsubsection{Classic VDB}
The Tukey post-hoc test results for the Classic VDB setup confirm that both HyDE and its combinations with Cohere Rerank and LLM Rerank significantly outperform the Naive RAG, aligning with the boxplot observations of higher retrieval precision. Significant improvemnt is also observed for LLM Rerank alone. However, the Maximal Marginal Relevance (MMR) and Cohere Rerank do not show a significant improvement over Naive RAG, which is also reflected in their closer median precision scores in the boxplots. Interestingly, the Multi Query + Reciprocal technique, while statistically significant, presents a mean difference suggesting lower performance compared to Naive RAG, contradicting the anticipated outcome and calling for additional scrutiny.
\begin{table}[ht]
\centering
\small
\begin{tabular}{llccc}
\hline
\textbf{Technique} & \textbf{Comparison} & \textbf{Mean Diff.} & \textbf{P-adj} & \textbf{Reject Null} \\
\hline
Cohere Rerank & Naive RAG & -0.0150 & 0.4515 & False \\
HyDE & Naive RAG & -0.0648 & 0.0000 & True \\
HyDE + Cohere Rerank & Naive RAG & -0.0371 & 0.0000 & True \\
HyDE + LLM Rerank & Naive RAG & -0.0749 & 0.0000 & True \\
LLM Rerank & Naive RAG & -0.0514 & 0.0000 & True \\
Maximal Marginal Relevance (MMR) & Naive RAG & -0.0156 & 0.3787 & False \\
Multi Query + Cohere Rerank & Naive RAG & 0.0012 & 1.0000 & False \\
Multi Query + Reciprocal & Naive RAG & 0.0542 & 0.0000 & True \\
\hline
\end{tabular}
\captionsetup{width=.8\linewidth, font=footnotesize} 
\caption{Tukey's HSD test results comparing RAG techniques to Naive RAG, all within the Classic VDB framework}
\label{tab:tukey-results}
\end{table}

Next, the focus is on techniques which have shown statistically significant improvement over the Naive RAG approach. The table below presents the results from Tukey's post-hoc tests, contrasting each of the high-performing techniques against each other.
\begin{table}[ht]
\centering
\small
\begin{tabular}{llccc}
\hline
\textbf{Technique} & \textbf{Comparison} & \textbf{Mean Diff.} & \textbf{P-adj} & \textbf{Reject Null} \\
\hline
HyDE & HyDE + Cohere Rerank & -0.0277 & 0.0002 & True \\
HyDE & HyDE + LLM Rerank & 0.0101 & 0.3255 & False \\
HyDE & LLM Rerank & -0.0134 & 0.1175 & False \\
HyDE + Cohere Rerank & HyDE + LLM Rerank & 0.0378 & 0.0000 & True \\
HyDE + Cohere Rerank & LLM Rerank & 0.0143 & 0.0842 & False \\
HyDE + LLM Rerank & LLM Rerank & -0.0235 & 0.0015 & True \\
\hline
\end{tabular}
\captionsetup{width=.8\linewidth, font=footnotesize} 
\caption{{Tukey's HSD test results of RAG techniques that offer significant improvement over Naive RAG}}
\label{tab:pairwise-tukey}
\end{table}

The combination of HyDE and LLM Rerank emerged as the most potent in enhancing retrieval precision within the Classic VDB framework, surpassing other techniques. However, this superior performance comes with higher latency and cost implications due to the additional LLM calls required for both reranking and hypothetical document embedding. Close second is HyDE alone, not showing significant difference from HyDE + LLM rerank combination. Experiments including Cohere Rerank did not demonstrate anticipated benefits.

\subsubsection{Sentence window}
This section delves into the analysis of Sentence Window retrieval techniques. First, the worst sentence window retriever technique is compared with the best classic VDB technique. 

\begin{table}[ht]
\centering
\small
\begin{tabular}{llccc}
\hline
\textbf{Technique} & \textbf{Comparison} & \textbf{Mean Diff.} & \textbf{P-adj} & \textbf{Reject Null} \\
\hline
HyDE + LLM Rerank & Sentence Window + Cohere Rerank & 0.1021 & 0.0000 & True \\
\hline
\end{tabular}
\captionsetup{width=.8\linewidth, font=footnotesize} 
\caption{Tukey's HSD test results for the best performing Classic VDB technique against the worst Sentence Window retrieval technique.}
\label{tab:best-classic-vdb-vs-sentence-window}
\end{table}

The Tukey's HSD test clearly indicates that even the least performing Sentence Window technique surpasses the best Classic VDB method in retrieval precision. This underscores the potential of the Sentence Window approach in RAG systems. However, the contrasting results from the Answer Similarity metric serve as a reminder to interpret these findings cautiously.  
\newline

Next is a comparison of individual Sentence Window retrieval variants.

\begin{table}[ht]
\centering
\small
\begin{tabular}{llccc}
\hline
\textbf{Base Technique} & \textbf{Compared Technique} & \textbf{Mean Diff.} & \textbf{P-adj} & \textbf{Reject Null} \\
\hline
Sentence Window & Sentence Window + Cohere Rerank & 0.0090 & 0.9768 & False \\
Sentence Window & Sentence Window + HyDE & -0.0025 & 1.0000 & False \\
Sentence Window & Sentence Window + HyDE + Cohere Rerank & -0.0078 & 0.9945 & False \\
Sentence Window & Sentence Window + LLM Rerank & 0.0332 & 0.0000 & True \\
\hline
\end{tabular}
\captionsetup{width=1\linewidth, font=footnotesize} 
\caption{Tukey's HSD test results for Sentence Window retrieval enhancements}
\label{tab:sentence-window-enhancements}
\end{table}

The Tukey's HSD test delineates Sentence Window retrieval with LLM Rerank as the only variant to offer a statistically significant improvement over the base Sentence Window technique.

\subsubsection{Document Summary Index}
The Document Summary Index technique was analyzed, focusing on two variations: one augmented with Cohere Rerank and another with HyDE plus Cohere Rerank. The choice to limit the study to these two is due to computational constraints and the need for comparability across experiments. The table below shows that there is no significant difference between the two techniques. 

\begin{table}[ht]
\centering
\small
\begin{tabular}{llccc}
\hline
\textbf{Technique} & \textbf{Comparison} & \textbf{Mean Diff.} & \textbf{P-adj} & \textbf{Reject Null} \\
\hline
Doc Sum + Cohere Re & Doc Summ + HyDE + Cohere Re & 0.0109 & 0.8935 & False \\
\hline
\end{tabular}
\captionsetup{width=1\linewidth, font=footnotesize} 
\caption{Tukey's HSD test results comparing Document Summary Index variants}
\label{tab:doc-summary-comparison}
\end{table}

To finish off the analysis, a basic version of every vector database set up were compared with another, i.e. Sentence Window, Naive RAG and Document Summary with Cohere rerank. Utilizing plain Document Summary without enhancements was not feasible for this analysis, as it aggregates multiple chunks into one summary, leading to results not directly comparable to other techniques that operate on different chunk quantities. 

\begin{table}[H]
\centering
\small
\begin{tabular}{llccc}
\hline
\textbf{Technique} & \textbf{Comparison} & \textbf{Mean Diff.} & \textbf{P-adj} & \textbf{Reject Null} \\
\hline
Doc Summ Index + Cohere Rerank & Classic VDB + Naive RAG & 0.0545 & 0.0000 & True \\
Doc Summ Index + Cohere Rerank & Sentence Window Retrieval & 0.1679 & 0.0000 & True \\
Classic VDB + Naive RAG & Sentence Window Retrieval & 0.1134 & 0.0000 & True \\
\hline
\end{tabular}
\captionsetup{width=1\linewidth, font=footnotesize} 
\caption{Tukey's HSD test results comparing the performance of Sentence Window retrieval and Document Summary Index + Cohere Rerank against the baseline Naive RAG}
\label{tab:tukey-comparison-enhancements}
\end{table}

The Tukey's HSD test results establish the Sentence Window retrieval as the leading technique, surpassing the Document Summary Index in precision. Document Summary Index with Cohere Rerank trails behind as a viable second, whereas the Classic VDB, in its standard form, demonstrates the least retrieval precision among the evaluated techniques.
\newline

\section{Limitations}
\begin{itemize}
   \item \textbf{Model selection:} We used GPT-3.5-turbo for evaluating responses due to the constraints of Tonic Validate, which requires the use of OpenAI models. The choice of GPT-3.5-turbo, while cost-effective, may not offer the same depth of analysis as more advanced models like GPT-4.
   \item \textbf{Data and question scope:} The study was conducted using a singular dataset and a set of 107 questions, which may affect the generalizability of the findings across different LLM applications. Expanding the variety of datasets and questions could potentially yield more comprehensive insights.
   \item \textbf{Chunking variability:} While the use of multiple chunking strategies allowed for a comprehensive evaluation of different retrieval methods, it also highlighted the inherent challenges in directly comparing their performance against the same metrics. Each retrieval method required a distinct chunking approach tailored to its specific needs. For instance, the sentence window retrieval method necessitated overlapping chunks of consecutive sentences, while the document summary index used larger chunks to leverage the language model's summarization capabilities effectively. Consequently, the retrieval methods were evaluated on chunk types with varying degrees of context and information density, making it difficult to draw definitive conclusions about their relative strengths and weaknesses. This limitation stems from the fundamental differences in how these retrieval methods operate and the distinct chunking requirements they impose. 
   \item \textbf{Evaluation metrics:} The lack of a clear consensus on the optimal metrics for evaluating RAG systems means our chosen metrics—Retrieval Precision and Answer Similarity—are based on conceptual alignment rather than empirical evidence of their efficacy. This highlights an area for future research to solidify the evaluation framework for RAG systems.
   \item \textbf{Technique Selection:} The subset of RAG techniques evaluated, while selected based on current relevance and potential, is not exhaustive. Excluded techniques such as Step back prompting \citep{dai2023stepback}, Auto-merging retrieval \citep{phaneendra2023automergeretrieval}, and Hybrid search \citep{akash2023hybridsearch} reflect the study's scope limitation and the subjective nature of selection. Future research should consider these and other emerging methods to broaden the understanding of RAG system enhancements.
\end{itemize}

\section{Conclusion}
Our investigation into Retrieval-Augmented Generation (RAG) techniques has identified HyDE and LLM reranking as notable enhancers of retrieval precision in LLMs. These approaches, however, necessitate additional LLM queries, incurring greater latency and cost. Surprisingly, established techniques like MMR and Cohere rerank did not demonstrate significant benefits, and Multi-query was found to be less effective than baseline Naive RAG.

The results demonstrate the efficacy of the Sentence Window Retrieval technique in achieving high precision for retrieval tasks, although a discrepancy was observed between retrieval precision and answer similarity scores. Given its conceptual similarity to Sentence Window retrieval, we suggest that Auto-merging retrieval \citep{phaneendra2023automergeretrieval} might offer comparable benefits, warranting future investigation. The Document Summary Index approach also exhibited satisfactory performance, however, it requires an upfront investment in generating summaries for each document in the corpus.

Due to constraints such as dataset singularity, limited questions, and the use of GPT-3.5-turbo for evaluation, the results may not fully capture the potential of more advanced models. Future studies with broader datasets and higher-capability LLMs could provide more comprehensive insights. This research contributes a foundational perspective to the field, encouraging subsequent works to refine, validate, and expand upon our findings.

To facilitate this continuation of research and allow for the replication and extension of our work, we have made our experimental pipeline available through a publicly accessible GitHub repository. \citep{ARAGOG2024}

\section{Future Work}
\begin{itemize}
   \item \textbf{Knowledge Graph RAG:} Integrating Knowledge Graphs (KGs) with RAG systems represents a promising direction for enhancing retrieval precision and contextual relevance. KGs offer a well-organized framework of relationship-rich data that could refine the retrieval phase of RAG systems \citep{neo4j2023knowledgegraph}. Although setting up such systems is resource-demanding, the potential for significantly improved retrieval processes justifies further investigation.
    \item \textbf{Unfrozen RAG systems:} Unlike the static application of RAG systems in our study, future investigations can benefit from adapting RAG components, including embedding models and rerankers, directly to specific datasets \citep{gao2024retrievalaugmented, kiela2024unfrozenrag}. This "unfrozen" approach allows for fine-tuning on nuanced use-case data, potentially enhancing system specificity and output quality. Exploring these adaptations could lead to more adaptable and effective RAG systems tailored to diverse application needs.
    \item \textbf{Experiment replication across diverse datasets:} To ensure the robustness and generalizability of our findings, it is imperative for future research to replicate our experiments using a variety of datasets. Conducting these experiments across multiple datasets is important to verify the applicability of our results and to identify any context-specific adjustments needed.
    \item \textbf{Auto-RAG:} The idea of automatically optimizing RAG systems, akin to Auto-ML's approach in traditional machine learning, presents a significant opportunity for future exploration. Currently, selecting the optimal configuration of RAG components — e.g., chunking strategies, window sizes, and parameters within rerankers — relies on manual experimentation and intuition. An automated system could systematically explore a vast space of RAG configurations and select the very best model \citep{AutoRAG2024}.
\end{itemize}

\bibliography{references}

\begin{thebibliography}{21}
\providecommand{\natexlab}[1]{#1}
\providecommand{\url}[1]{\texttt{#1}}
\expandafter\ifx\csname urlstyle\endcsname\relax
  \providecommand{\doi}[1]{doi: #1}\else
  \providecommand{\doi}{doi: \begingroup \urlstyle{rm}\Url}\fi

\bibitem[Akash(2023)]{akash2023hybridsearch}
Akash.
\newblock Hybrid search: Optimizing rag implementation.
\newblock \url{https://medium.com/@csakash03/hybrid-search-is-a-method-to-optimize-rag-implementation-98d9d0911341}, 2023.
\newblock Accessed: 2024-04-01.

\bibitem[Bratanic(2023)]{neo4j2023knowledgegraph}
T.~Bratanic.
\newblock Using a knowledge graph to implement a rag application.
\newblock \url{https://neo4j.com/developer-blog/knowledge-graph-rag-application/}, 2023.
\newblock Accessed: 2024-03-24.

\bibitem[Carbonell and Goldstein(1998)]{carbonell1998usemmr}
J.~Carbonell and J.~Goldstein.
\newblock The use of mmr, diversity-based reranking for reordering documents and producing summaries.
\newblock \url{https://www.cs.cmu.edu/~jgc/publication/The_Use_MMR_Diversity_Based_LTMIR_1998.pdf}, 1998.
\newblock Accessed: 2024-03-24.

\bibitem[Dai et~al.(2023)Dai, Callan, Chang, Chen, Guu, Han, Hashimoto, He, Joshi, Jurafsky, Karishnamurthy, Khashabi, Kiela, Kumar, Lan, Lewis, Ma, Min, Neelakantan, Ng, Pasupat, Qi, Raffel, Roller, Shih, and Zettlemoyer]{dai2023stepback}
Z.~Dai, J.~Callan, K.-W. Chang, D.~Chen, K.~Guu, X.~Han, K.~Hashimoto, H.~He, M.~Joshi, D.~Jurafsky, J.~Karishnamurthy, D.~Khashabi, D.~Kiela, A.~Kumar, Z.~Lan, M.~Lewis, X.~Ma, S.~Min, A.~Neelakantan, A.~Y. Ng, P.~Pasupat, P.~Qi, C.~Raffel, S.~Roller, K.~Shih, and L.~Zettlemoyer.
\newblock Step back prompting: Enhancing llms with historical context retrieval.
\newblock \url{https://arxiv.org/abs/2310.06117}, 2023.

\bibitem[Devlin et~al.(2019)Devlin, Chang, Lee, and Toutanova]{devlin2019bert}
J.~Devlin, M.-W. Chang, K.~Lee, and K.~Toutanova.
\newblock Bert: Pre-training of deep bidirectional transformers for language understanding, 2019.

\bibitem[Gao et~al.(2022)Gao, Ma, Lin, and Callan]{gao2022precise}
L.~Gao, X.~Ma, J.~Lin, and J.~Callan.
\newblock Precise zero-shot dense retrieval without relevance labels, 2022.

\bibitem[Gao et~al.(2024)Gao, Xiong, Gao, Jia, Pan, Bi, Dai, Sun, Guo, Wang, and Wang]{gao2024retrievalaugmented}
Y.~Gao, Y.~Xiong, X.~Gao, K.~Jia, J.~Pan, Y.~Bi, Y.~Dai, J.~Sun, Q.~Guo, M.~Wang, and H.~Wang.
\newblock Retrieval-augmented generation for large language models: A survey, 2024.

\bibitem[{James Calam}(2023)]{calam2023aiarxiv}
{James Calam}.
\newblock Ai arxiv dataset.
\newblock \url{https://huggingface.co/datasets/jamescalam/ai-arxiv}, 2023.
\newblock Accessed: 2024-03-24.

\bibitem[Jiang et~al.(2023)Jiang, Xu, Gao, Sun, Liu, Dwivedi-Yu, Yang, Callan, and Neubig]{jiang2023active}
Z.~Jiang, F.~F. Xu, L.~Gao, Z.~Sun, Q.~Liu, J.~Dwivedi-Yu, Y.~Yang, J.~Callan, and G.~Neubig.
\newblock Active retrieval augmented generation, 2023.

\bibitem[Kiela(2024)]{kiela2024unfrozenrag}
D.~Kiela.
\newblock Stanford cs25: V3 i retrieval augmented language models.
\newblock \url{https://www.youtube.com/watch?v=mE7IDf2SmJg}, 2024.
\newblock Accessed: 2024-03-24.

\bibitem[Langchain(2023)]{langchain_blog2023query}
Langchain.
\newblock Query transformations.
\newblock \url{https://blog.langchain.dev/query-transformations/}, 2023.
\newblock Accessed: 2024-03-23.

\bibitem[Liu(2023{\natexlab{a}})]{llamaindex2023docsummary}
J.~Liu.
\newblock A new document summary index for llm-powered qa systems.
\newblock \url{https://www.llamaindex.ai/blog/a-new-document-summary-index-for-llm-powered-qa-systems-9a32ece2f9ec}, 2023{\natexlab{a}}.
\newblock Accessed: 2024-03-23.

\bibitem[Liu(2023{\natexlab{b}})]{llamaindex2023llmrerank}
J.~Liu.
\newblock Using llms for retrieval and reranking.
\newblock \url{https://www.llamaindex.ai/blog/using-llms-for-retrieval-and-reranking-23cf2d3a14b6}, 2023{\natexlab{b}}.
\newblock Accessed: 2024-03-24.

\bibitem[Liu et~al.(2019)Liu, Ott, Goyal, Du, Joshi, Chen, Levy, Lewis, Zettlemoyer, and Stoyanov]{liu2019roberta}
Y.~Liu, M.~Ott, N.~Goyal, J.~Du, M.~Joshi, D.~Chen, O.~Levy, M.~Lewis, L.~Zettlemoyer, and V.~Stoyanov.
\newblock Roberta: A robustly optimized bert pretraining approach, 2019.

\bibitem[{Markr.AI}(2024)]{AutoRAG2024}
{Markr.AI}.
\newblock Autorag: A framework for automated retrieval-augmented generation.
\newblock \url{https://github.com/Marker-Inc-Korea/AutoRAG}, 2024.
\newblock Accessed: 2024-03-24.

\bibitem[Phaneendra(2023)]{phaneendra2023automergeretrieval}
K.~Phaneendra.
\newblock Deep dive into advanced rag applications in llm-based systems.
\newblock \url{https://phaneendrakn.medium.com/deep-dive-into-advanced-rag-applications-in-llm-based-systems-1ccee0473b3b}, 2023.
\newblock Accessed: 2024-04-01.

\bibitem[Pinecone(2023)]{pinecone2023rerankers}
Pinecone.
\newblock Rerankers.
\newblock \url{https://www.pinecone.io/learn/series/rag/rerankers/}, 2023.
\newblock Accessed: 2024-03-24.

\bibitem[{Predlico}(2024)]{ARAGOG2024}
{Predlico}.
\newblock Aragog - advanced retrieval augmented generation output grading.
\newblock \url{https://github.com/predlico/ARAGOG}, 2024.
\newblock Accessed: 2024-03-24.

\bibitem[{RAGAS Documentation}(2023)]{ragas2023metrics}
{RAGAS Documentation}.
\newblock Metrics.
\newblock \url{https://docs.ragas.io/en/v0.0.17/concepts/metrics/index.html}, 2023.
\newblock Accessed: 2024-03-24.

\bibitem[{Tonic AI}(2023)]{tonic2023ragmetrics}
{Tonic AI}.
\newblock About rag metrics: Tonic validate rag metrics summary.
\newblock \url{https://docs.tonic.ai/validate/about-rag-metrics/tonic-validate-rag-metrics-summary}, 2023.
\newblock Accessed: 2024-03-24.

\bibitem[Yang(2023)]{towardsdatascience2023advancedrag}
S.~Yang.
\newblock Advanced rag 01: Small to big retrieval.
\newblock \url{https://towardsdatascience.com/advanced-rag-01-small-to-big-retrieval-172181b396d4}, 2023.
\newblock Accessed: 2024-03-23.

\end{thebibliography}

\end{document}